\documentclass[runningheads]{llncs}
\usepackage[T1]{fontenc}
\usepackage{xcolor}
\usepackage{float} 
\usepackage{graphicx}
\usepackage{multirow}
\begin{document}
\title{Text Reading Order in Uncontrolled Conditions by Sparse Graph Segmentation}
\titlerunning{Text Reading Order by Graph Segmentation}
\author{Renshen Wang \and
Yasuhisa Fujii \and
Alessandro Bissacco}
\authorrunning{ }
\institute{Google Research\\
 \email{\{rewang, yasuhisaf, bissacco\}@google.com}
}

\maketitle              % typeset the header of the contribution
\begin{abstract}
Text reading order is a crucial aspect in the output of an OCR engine, with a large impact on downstream tasks. Its difficulty lies in the large variation of domain specific layout structures, and is further exacerbated by real-world image degradations such as perspective distortions. We propose a lightweight, scalable and generalizable approach to identify text reading order with a multi-modal, multi-task graph convolutional network (GCN) running on a sparse layout based graph. Predictions from the model provide hints of bidimensional relations among text lines and layout region structures, upon which a post-processing cluster-and-sort algorithm generates an ordered sequence of all the text lines. The model is language-agnostic and runs effectively across multi-language datasets that contain various types of images taken in uncontrolled conditions, and it is small enough to be deployed on virtually any platform including mobile devices.

\end{abstract}
\section{Introduction}
\label{sec:introduction}

Optical character recognition (OCR) technology has been developed to extract text reliably from various types of image sources \cite{photoocr2013}. Key components of an OCR system include text detection, recognition and layout analysis. As machine learning based digital image processing systems are nowadays ubiquitous and widely applied, OCR has become a crucial first step in the pipeline to provide text input for downstream tasks such as information extraction, text selection and screen reading.

Naturally, most image-to-text applications require very accurate OCR results to work well. This requirement is not only on text recognition --- reading order among the recognized text lines is almost always as important as the recognition quality. The reason is self-evident for text selection (copy-paste) and text-to-speech tasks. And for structured document understanding like LayoutLM \cite{layoutlmv2}, DocFormer \cite{docformer2021}, FormNet \cite{formnet2022}, etc., the order of the input text also has a profound effect as most of these models have positional encoding attached to input text features, and a sequential labeling task for output. Input text order can sometimes be the key factor for the successful extraction of certain entities. % --- e.g. multi-line address entities that can be broken by a wrong reading order.

\begin{figure}[t]
\includegraphics[height=\textwidth,angle=90]{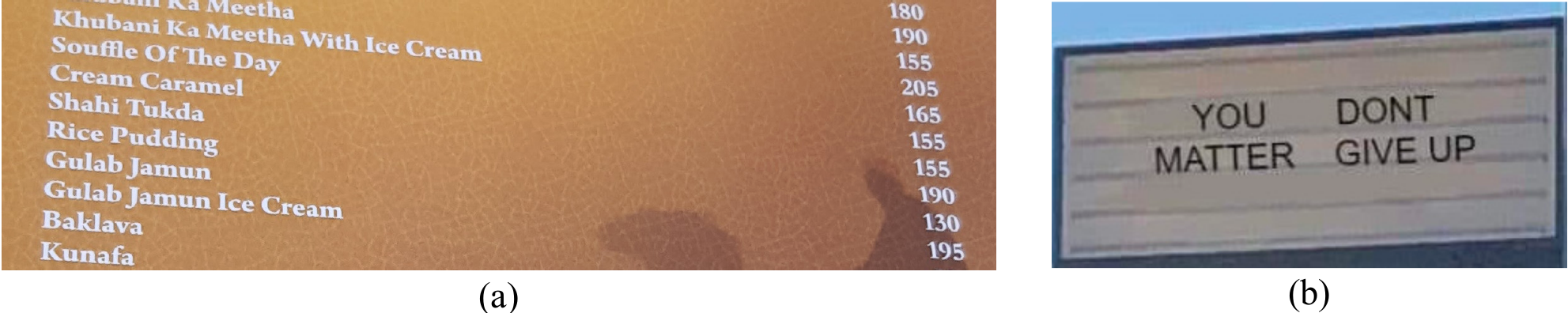}
\vspace{-7mm}
\centering
\caption{Hard examples for text reading order. (a) A cropped image of a menu with dish names and prices, where a correct reading order necessarily needs correct association between each dish name and its price, which is a hard task for humans without the full image context due to the perspective distortion in the image. (b) A text layout intentionally made to have two different reading order interpretations, both valid, but with completely opposite meanings.}
\label{fig:1}       % Give a unique label
\end{figure}

Depending on the text layout, the difficulty of deciding its reading order varies greatly. It can be as simple as sorting all the text lines by y-coordinates, but can also be hard like the images in Figure \ref{fig:1}. Even if we exclude corner cases like these, there are still complexities brought by the diversity of layout structures which are often domain specific. Previous studies have tackled the problem in different ways. Rule based approaches like \cite{bidim2003,xycut2005,argmentation2014} usually aim at one specific domain, while learning based approaches like \cite{datamining2007,e2esequence2020,layoutreader2021} are more general but have scalability issues (more discussions in the following section).

\begin{figure}[t]
\includegraphics[height=0.9\textwidth,angle=90]{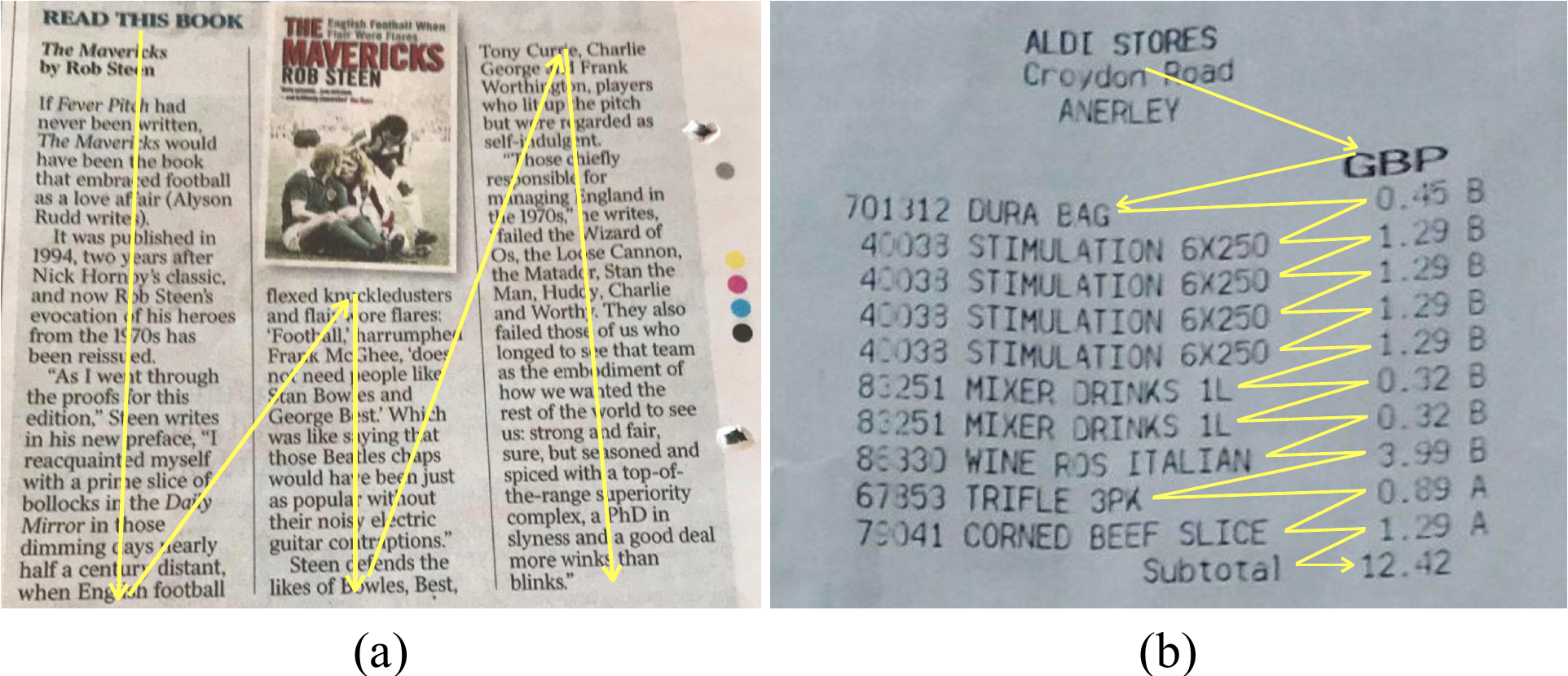}
\vspace{-3mm}
\centering
\caption{Two major patterns of reading order. (a) Column-wise order, most common in printed media like newspapers and magazines. (b) Row-wise order, usually in receipts, forms and tabular text blocks.}
\label{fig:2}       % Give a unique label
\vspace{-3mm}
\end{figure}

In this paper, we propose a composite method that uses both machine learning model and rule based sorting to achieve best results. It is based on the observation from \cite{bidim2003} that most reading order sequences are in one of the two patterns --- column-wise and row-wise  --- as illustrated in Figure \ref{fig:2}.

We use a graph convolutional network that takes spatial-image features from the input layout and image, and segments the layout into two types of regions where the paragraphs can be properly sorted by the type of their patterns. A $\beta$-skeleton graph built on boxes \cite{par2022} enables efficient graph convolutions while also providing edge bounding boxes for RoI (regions of interest) pooling from the image feature map. A post-processing cluster-and-sort algorithm finalizes the overall reading order based on model predictions. This unique combination gives us an effective, lightweight, scalable and generalizable reading order solution.

\section{Related Work}
\label{sec:related}

Two types of related work are discussed in this section. The first subsection includes previous reading order efforts, and the second subsection discusses other multi-modal image-text-spatial models that share some of the components with our approach.

\subsection{Reading Order Detection}
\label{sec:related_reading_order}

Previous studies have tackled the reading order problem in various ways. We roughly categorize them into rule based sorting \cite{hplayout2003,bidim2003,xycut2005,argmentation2014} and machine-learning based sequence prediction \cite{datamining2007,e2esequence2020,layoutreader2021,readingorderhand2022}, etc.

Topological sort was proposed in \cite{hplayout2003} for document layout analysis where partial orders are based on x/y interval overlaps among text lines. It can produce reading order patterns like Figure \ref{fig:2} (a) for multi-column text layouts. A bidimensional relation rule proposed in \cite{bidim2003} provides similar topological rules, and in addition provides a row-wise rule by inverting the x/y axes from column-wise. An argumentation based approach in \cite{argmentation2014} works on similar rules derived from text block relations. For large text layout with hierarchies, XY-Cut \cite{xycut2005,xylayoutlm2022} can be an effective way for some layout types to order all the text blocks top-to-bottom and left-to-right. These rule based approaches can work accurately for documents in certain domains. But without extra signals, they will fail for out-of-domain cases like Figure \ref{fig:2} (b).

Machine learning based approaches are designed to learn from training examples across different domains to enable a general solution. The data mining approach in \cite{datamining2007} learns partial order among text blocks from their spatial features and identifies reading order chains from the partial orders. A similar approach in \cite{readingorderhand2022} trains a model to predict pairwise order relations among text regions and curves for handwritten documents. The major limitation is that any partial order between two entities are derived from their own spatial features without the layout structure information in their neighborhood. So these models may not be able to identify the layout structure among a group of text lines and therefore fail to find the correct pattern.

Graph convolutional networks and transformer models provide mechanisms for layout-aware signals by interactions between layout entities. A text reorganization model introduced in \cite{e2esequence2020} uses a graph convolutional encoder and a pointer network decoder to reorder text blocks. With a fully-connected graph at its input, the graph encoder functions similarly as a transformer encoder. Image features are added to graph nodes by RoI pooling on node boxes with bi-linear interpolation. Another work LayoutReader \cite{layoutreader2021} uses a transformer based architecture on spatial-text features instead of spatial-image features to predict reading order sequence on words. The text features enable it to use the powerful LayoutLM \cite{layoutlm} model, but also make it less generalizable. These models are capable of predicting reading order within complex layout structures. However, there are scalability issues in two aspects:

\begin{itemize}
  \item Run time scales quadratically with input size. Whether in the graph convolutional encoder with full connections or the sequence pointer decoder, most of the components have $O(n^2)$ time complexity, and may become too slow for applications with dense text.
  
  \item Accuracy scales inversely with input size. The fully-connected self-attention mechanism in the encoder takes all the text entities to calculate a global attention map, which introduces noises to the reading order signals that should be decidable from local layout structures. The sequence decoder uses softmax probabilities to determine the output index for each step, where the output range increases with input size, and so does the chance of errors. Figure \ref{fig:11} illustrates this limitation from our experiments. 
\end{itemize}

To summarize briefly, there are multiple effective ways to order OCR text by rule based or machine learning based methods, and in both categories there is room for improvement in generalizability and scalability.

\subsection{Spatial, Image Features and Multi-Modality}

Multi-modal transformer models have become mainstream for document or image understanding tasks. Related work include LayoutLM~\cite{layoutlm,layoutlmv2,layoutlmv3,xylayoutlm2022}, DocFormer~\cite{docformer2021}, SelfDoc~\cite{selfdoc2021}, UDoc~\cite{unidoc2021}, StrucText~\cite{structext2021}, TILT~\cite{tilt2021}, LiLT~\cite{lilt2022}, FormNet \cite{formnet2022}, PaLI \cite{pali}, etc.

Document image understanding starts with an OCR engine that provides text content as the main input for the language model. Alongside, the text bounding boxes associated with the words and lines provide important spatial features (sometimes called layout features or geometric features). Additionally, since not all visual signals are captured by the OCR engine, an image component in the model can help cover the extra contextual information from the input. Thus, a model to achieve best results should take all of the three available modalities.

For image features, most previous studies use RoI pooling \cite{roi2016} by the text bounding boxes from OCR, and the pooled features are attached to the corresponding text entity. It is effective for capturing text styles or colors, but less so for visual cues out of those bounding boxes, such as the curly separation lines in Figure \ref{fig:3}. While it is possible to use an image backbone with large receptive fields, like ResNet50 used in the UDoc model or U-Net used in the TILT model, it is not an ideal solution for two reasons:

\begin{itemize}
  \item In sparse documents, useful visual cues can be far from any text on the page.
  
  \item Large receptive fields bring in extra noise from regions irrelevant to the features we need.
\end{itemize}

Thus, it will be more effective to have image RoI boxes that cover pairs of text bounding boxes. A sparse graph like $\beta$-skeleton used in \cite{par2022} can provide the node pairs for such RoI pooling without significantly increasing the model's memory footprint and computational cost.

\section{Proposed Method}

Based on previous studies, we design a lightweight machine learning based approach with a model that is small in size, fast to run, and easy to generalize in uncontrolled conditions. 

\subsection{Strong Patterns of Reading Order}

From a set of real-world images annotated with reading order, we have an observation that matches very well with the bidimensional document encoding rules in \cite{bidim2003} --- column-wise text usually has a zigzag pattern of Figure \ref{fig:2} (a), and row-wise text has a similar but transposed zigzag like Figure \ref{fig:2} (b). Some images may contain both types of text, which makes the pattern more complex. But once the column-wise/row-wise type of a text region is decided, the reading order in this region mostly follows the pattern and can be determined with a topological sort according to the bidimensional rules. Figure \ref{fig:7} (a) shows an example of an annotated reading order sequence.

Based on this observation, learning text reading order becomes an image segmentation problem, as opposed to learning arbitrary global sequences of text entities. Instead of predicting the next entity in the entire image, we do a binary classification for each text entity on whether it's in a column-wise or row-wise pattern. Moreover, the pattern classification for a text line can be decided by local layout structures, and global attention maps are therefore unnecessary.

\subsection{Model Architecture}

\begin{figure}[t]
\includegraphics[height=0.7\textwidth,angle=90]{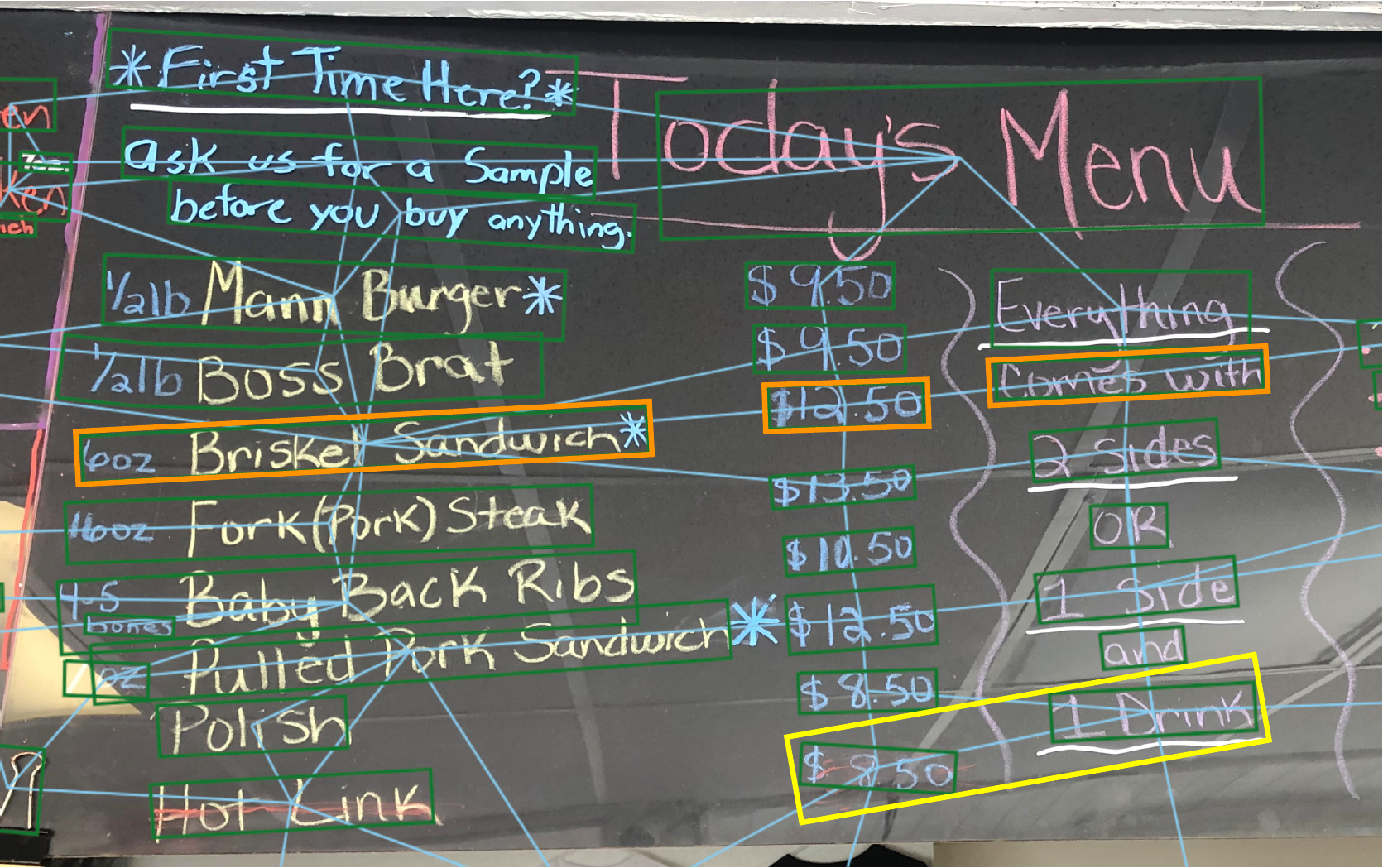}
\vspace{-2mm}
\centering
\caption{A cropped example of a $\beta$-skeleton graph \cite{par2022} constructed from text line boxes. Graph node boxes are shown in green and edge lines in cyan. The three orange colored text lines demonstrate how image features can help --- the 2nd and 3rd boxes are closer in distance, so spatial features may indicate they are in the same section, but the curly separation line between them indicates otherwise. The yellow box at the bottom is the minimum containing box of the two line boxes inside, where the RoI pooling can cover image features between these lines.}
\label{fig:3}       % Give a unique label
\end{figure}

We use a graph convolutional network (GCN) with a sparse graph construction because of the three major advantages listed here:

\begin{itemize}
  \item GCN models are equivariant to input order permutations. It is natural to assume that a model deciding reading order should not depend on the order of its input.
  
  \item With a sparse graph like $\beta$-skeleton, GCN computation scales linearly with input size. 
  
  \item Graph edges constructed from text boxes can provide edge bounding boxes, which are better for image feature RoI pooling (Figure \ref{fig:3}, Table \ref{tab:2}).
\end{itemize}

\begin{figure}[!b]
\vspace{5mm}
\includegraphics[height=\textwidth,angle=90]{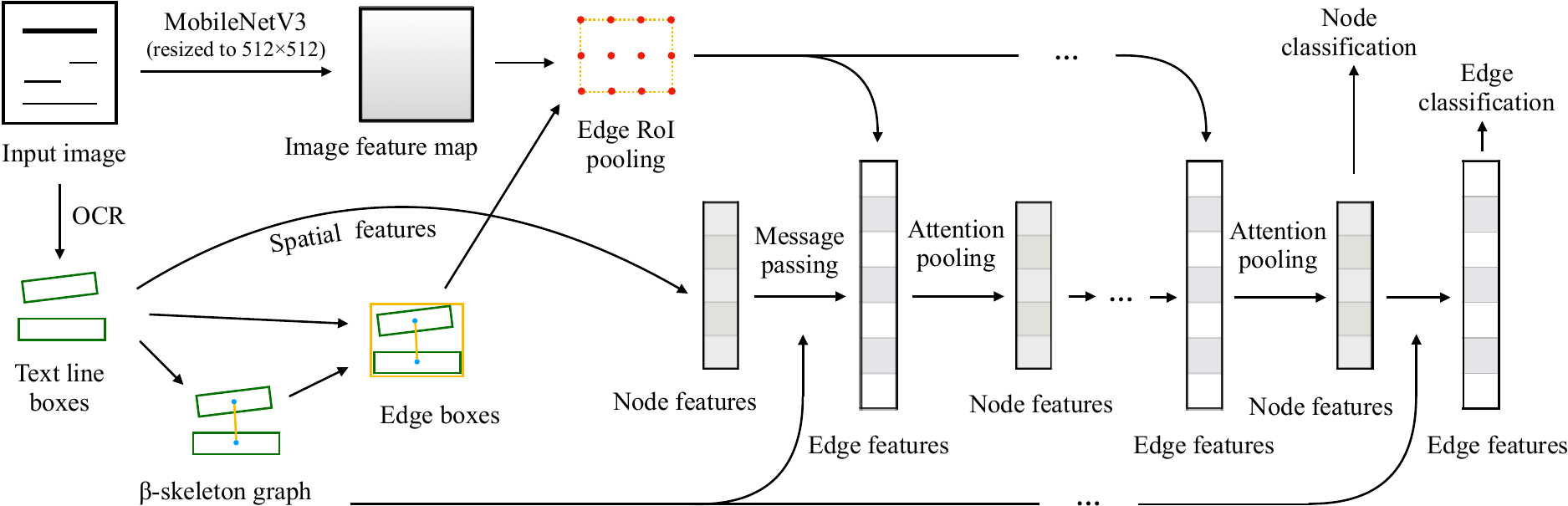}
\centering
\caption{Overview of the reading order multi-classifier model. Node classification predicts the reading order patterns, and edge classification predicts paragraph clustering. }
\label{fig:4}       % Give a unique label
\end{figure}

As illustrated in Figure \ref{fig:4}, we use an MPNN \cite{mpnn2017} variant of GCN as the main model backbone, and a $\beta$-skeleton graph \cite{beta1985} constructed with text line boxes as nodes. Similar configurations have been applied to other layout problems \cite{rope2021,par2022,unifiedgcn2022,formnet2022}, and graph construction details are available in \cite{par2022}. The main GCN input is from the spatial features of text line bounding boxes as node features, including $x$, $y$ coordinate values of the box corners, and the coordinate values multiplied by rotation angle coefficients $\cos \alpha$, $\sin \alpha$. The spatial features go through $T$ steps of graph convolution layers, each containing a node-to-edge ``message passing'' layer and edge-to-node aggregation layer with attention weighted pooling. 

\begin{figure}[t]
\includegraphics[height=0.7\textwidth,angle=-90]{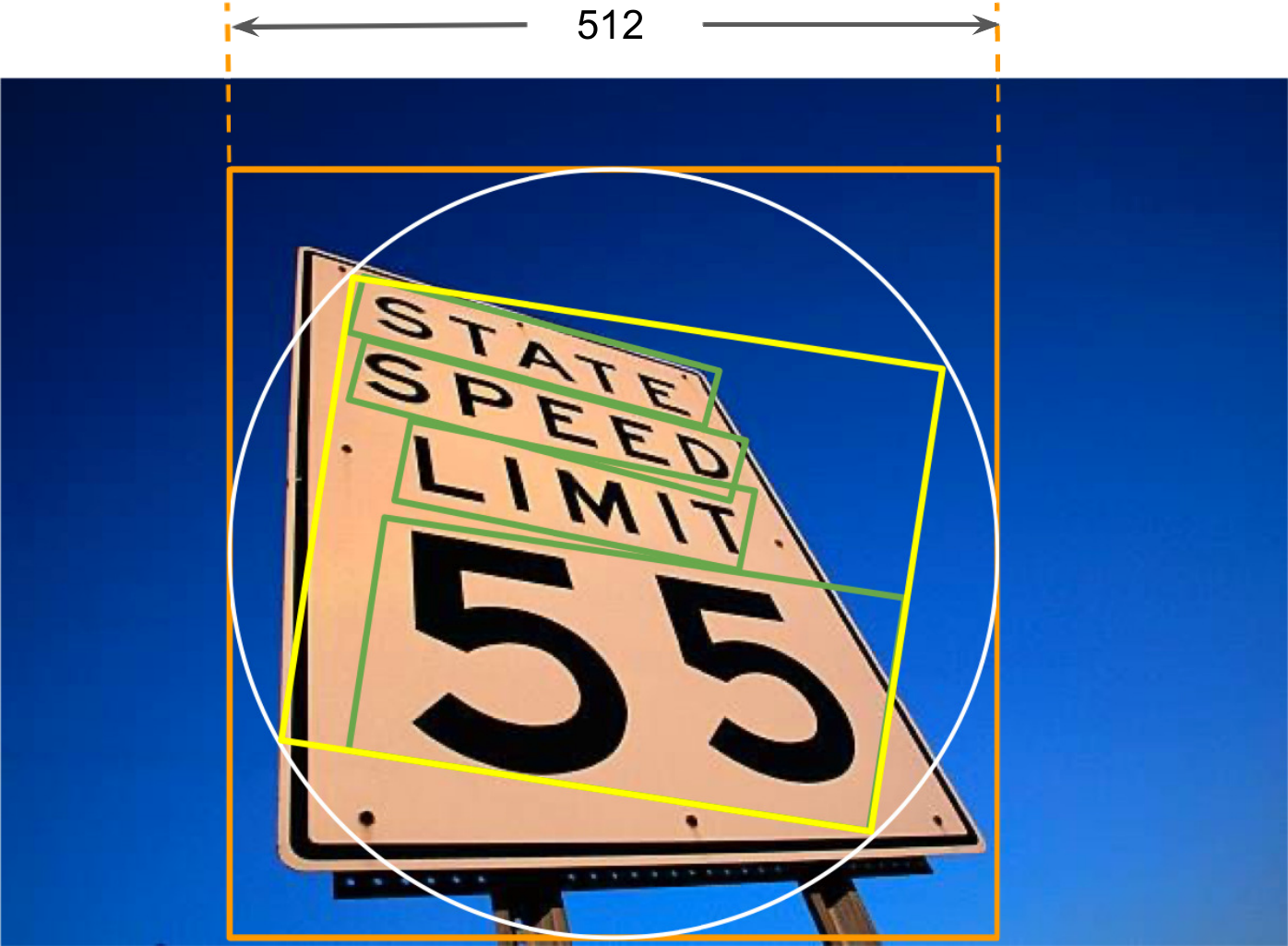}
\vspace{-1mm}
\centering
\caption{Image processing for the MobileNetV3 input. The inner yellow box is the minimum containing box of all the text lines in the image. If its diagonal $d$ is larger than 512, we scale down the image by $\frac{512}{d}$ so that all the text bounding boxes are contained in the white circle of diameter 512, and then we crop (maybe also pad) around this circle to get the final processed image. This process ensures that both the image and the layout can be randomly rotated during training without any line box moved out of boundary.}
\label{fig:5}       % Give a unique label
\end{figure}

Besides the main input from nodes, we add a side input of edge features from edge box RoI pooling on an image feature map to help capture potential visual cues surrounding text boxes. We use MobileNetV3-Small \cite{mobilenetv3} as the image backbone for its efficiency. Note that the purpose of this image backbone is not for a major task like object detection, but to look for auxiliary features like separation lines and color changes, so a small backbone is capable enough for our task. For the same reason, we reduce the MobileNetV3 input image size to 512$\times$512 to speed up training and inference. The details of the image processing are illustrated in Figure \ref{fig:5}. In most cases, the text content is no longer recognizable after such downsizing, but the auxiliary image features can be well preserved. We also make sure that the entire layout is contained in a circle of diameter 512 within the processed image, which enables random rotations during model training --- a key augmentation for our model to work in all conditions.

Language features are not included in order to keep the model minimal in size and independent of domain knowledge. Also, our annotated reading order data is limited in English only, upon which we try to train a universal model.

The GCN is a multi-task model that outputs both node and edge predictions. At node level, it predicts the reading order pattern on each line box (column-wise or row-wise). These predictions are essentially a segmentation for text regions where the lines can be sorted accordingly.

At edge level, the model predicts whether the two lines connected by an edge belong to the same paragraph. Thus, it works like the edge clustering models in \cite{par2022,unifiedgcn2022}, and we can improve the final reading order by grouping lines together within each paragraph. The reading order estimation by the grouping usually do not affect column-wise order among text lines, but can be critical in row-wise regions such as tables or forms with multi-line cells, e.g. Figure \ref{fig:10} (d).

It may be considered that a fully convolutional network can do similar segmentation tasks like \cite{segmentation2015,panoptic2019} on the input image. However, we have observed that such models are less effective for certain types of text content --- e.g. in Figure \ref{fig:2} (b), similar lines in the left column are grouped into a large paragraph, disrupting the row-wise reading order.

\subsection{Recovering Reading Order from Model Predictions}

With the $\beta$-skeleton graph that provides local connections among dense text boxes, the GCN model predicts on \textit{local} properties of the text, which can be aggregated to give us a \textit{global} reading order. To handle mixed column-wise and row-wise predictions as well as potential text rotations and distortions in the input image, we extend the rule based sorting in \cite{bidim2003,hplayout2003} and propose a hierarchical cluster-and-sort algorithm to recover the global reading order from line-level pattern predictions and clustered paragraphs. The following Algorithm 1 generates a set of clusters, each cluster $c_i$ contains a non-empty set of paragraphs and maybe a set of child clusters. Each cluster is also assigned a reading order pattern $R(c_i) \in \{\mathit{col}, \mathit{row}\}$, with $\mathit{col}$ for column-wise and $\mathit{row}$ for row-wise.

Row-wise text often involves sparse tables with components not directly connected by $\beta$-skeleton edges, so the hop edges like in \cite{unifiedgcn2022} can be helpful in step 4 of algorithm 1. More details can be added, e.g. setting an edge length threshold in step 3 to avoid merging distant clusters.

\begin{table}
\centering
\vspace{-5mm}
\noindent\begin{tabular}{p{0.9\textwidth}}
\noindent \hrulefill
\vspace{-3.5mm}

\noindent \hrulefill

\noindent \textbf{Algorithm 1: Hierarchical Clustering}

\noindent Input: Text line bounding boxes, $\beta$-skeleton graph $G$,

\hspace{9mm} GCN node predictions and edge predictions.

\vspace{-1mm}
\noindent \hrulefill
\vspace{-2mm}

\begin{enumerate}
\item Cluster lines into paragraphs $p_1, ..., p_n$ from edge predictions.

\item Each paragraph is initialized as a cluster, $c_i = \{p_i\}$. Reading order pattern $R(c_i)$ is the majority vote from the paragraph's line predictions .

\item For each edge $(i,j) \in G$, find cluster $c_a$ containing line $i$ and $c_b$ containing line $j$; if $R(c_a)=R(c_b)=\mathit{col}$, merge $c_a$ and $c_b$ into a bigger column-wise cluster.

\item For each edge $(i,j) \in G$ or hop edge $(i,j)$ ($\exists k$ that $(i,k) \in G$ and $(k,j) \in G$), find cluster $c_a$ containing line $i$ and $c_b$ containing line $j$; if $R(c_a)=R(c_b)=\mathit{row}$, merge $c_a$ and $c_b$ into a bigger row-wise cluster.

\item Calculate the containing box for each cluster. The rotation angle of the box is the circular mean angle of all the paragraphs in the cluster.

\item Sort the clusters by ascending area of their containing boxes.

\item For each cluster $c_i$, if its containing box $B(c_i)$ overlaps with $B(c_j)$ by area greater than $T \times Area(B(c_i))$, set $c_i$ as a child cluster of $c_j$.

\item Create a top level cluster with all the remaining clusters as its children.
\end{enumerate}
\vspace{-4mm}
\noindent \hrulefill
\end{tabular}
\vspace{-5mm}
\end{table}

Once the regions of reading order patterns are decided by the hierarchical clusters, we can use topological sort within each cluster as in Algorithm 2.

\begin{table}
\centering
\noindent\begin{tabular}{p{0.9\textwidth}}
\noindent \hrulefill
\vspace{-3.5mm}

\noindent \hrulefill

\noindent \textbf{Algorithm 2: Reading Order Sorting within a Cluster}

\noindent Input: Bounding boxes $b_1, ..., b_n$ from paragraphs or child

\hspace{9mm} clusters, the reading order pattern to sort with.

\vspace{-1mm}
\noindent \hrulefill
\vspace{-3mm}

\begin{enumerate}
\item Calculate $\alpha$, the circular mean angle from all the bounding box angles.

\item For each box $b_i$, rotate it around $(0, 0)$ by angle $-\alpha$.

\item For each box $b_i$, calculate its axis aligned minimum containing box $a_i$.

\item If the reading order pattern is column-wise,

\hspace{5mm} Add constraint $(i \rightarrow j)$ if $a_i$, $a_j$ overlap on x-axis and

\hspace{5mm} \hspace{36mm} $y_{center}(a_i) < y_{center}(a_j)$

\hspace{5mm} Sort $a_1, ..., a_n$ by ascending $x_{center}$

else {\color{gray} \hspace{17mm} $\#$ pattern is row-wise }

\hspace{5mm} Add constraint $(i \rightarrow j)$ if $a_i$, $a_j$ overlap on y-axis and

\hspace{5mm} \hspace{36mm} $x_{center}(a_i) < x_{center}(a_j)$

\hspace{5mm} Sort $a_1, ..., a_n$ by ascending $y_{center}$

\item Based on existing order, topologically sort $a_1, ..., a_n$ with the order constraints.

\end{enumerate}
\vspace{-4mm}
\noindent \hrulefill
\end{tabular}
\vspace{1mm}
\end{table}

\begin{figure}[!b]
\includegraphics[height=0.9\textwidth,angle=90]{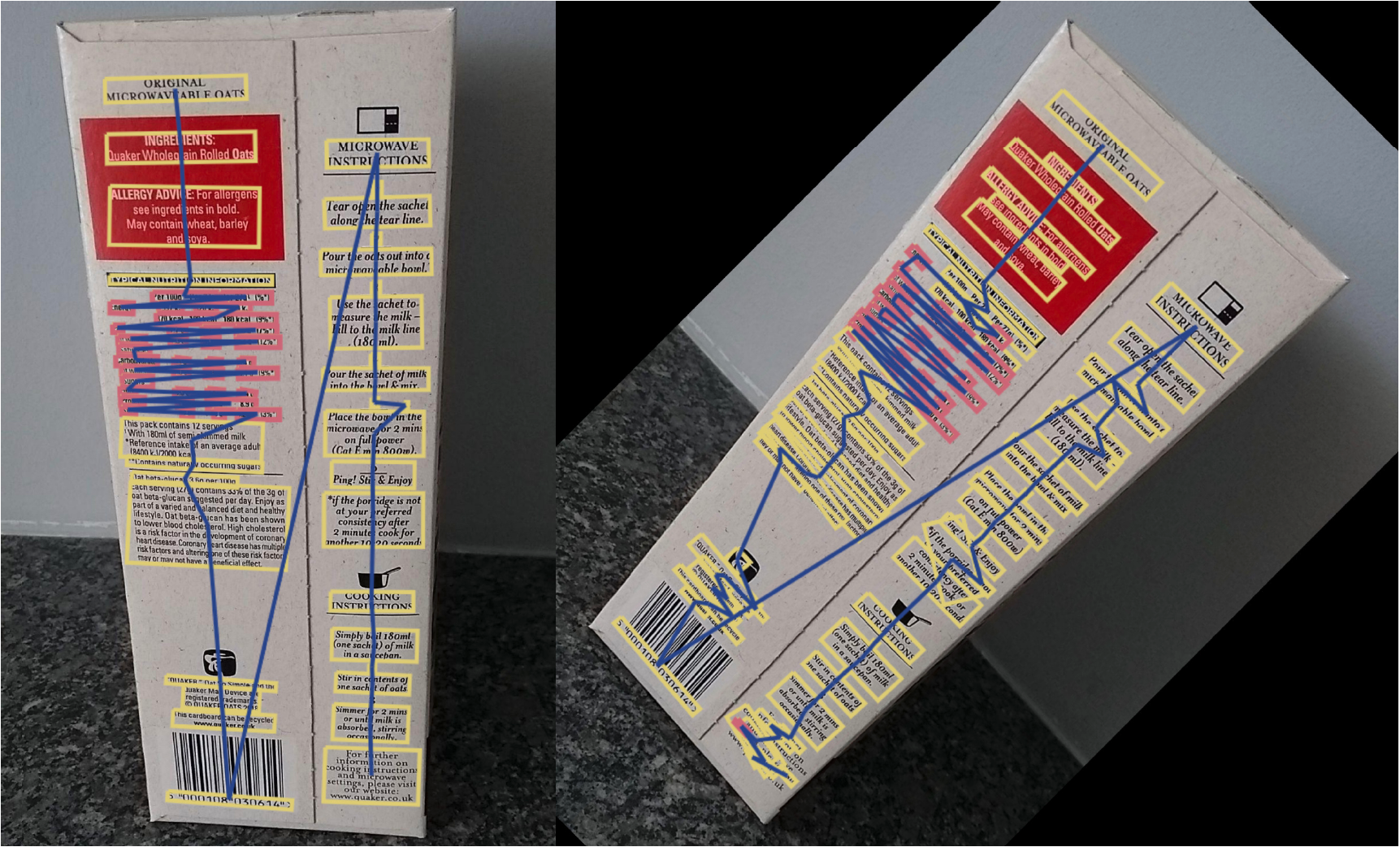}
\vspace{-2mm}
\centering
\caption{Reading order example at different angles. Paragraphs with column-wise pattern predictions are shown in yellow, row-wise in pink. The dark blue line shows the overall reading order among all paragraphs. }
\label{fig:6}       % Give a unique label
\end{figure}

With all the clusters sorted, an ordered traversal of the cluster hierarchy can give us the final reading order among all the paragraphs. Figure \ref{fig:6} shows the reading order on a packaging box at different camera angles. Note that the algorithms are not sensitive to bounding box angles, and the model is trained with randomly augmented data, so the rotation has minimal effect on the final result. It can even handle vertical text lines in Chinese/Japanese with the vertical lines regarded as rotated horizontal lines.

\subsection{Data Labeling}

\begin{figure}[t]
\includegraphics[height=0.85\textwidth,angle=90]{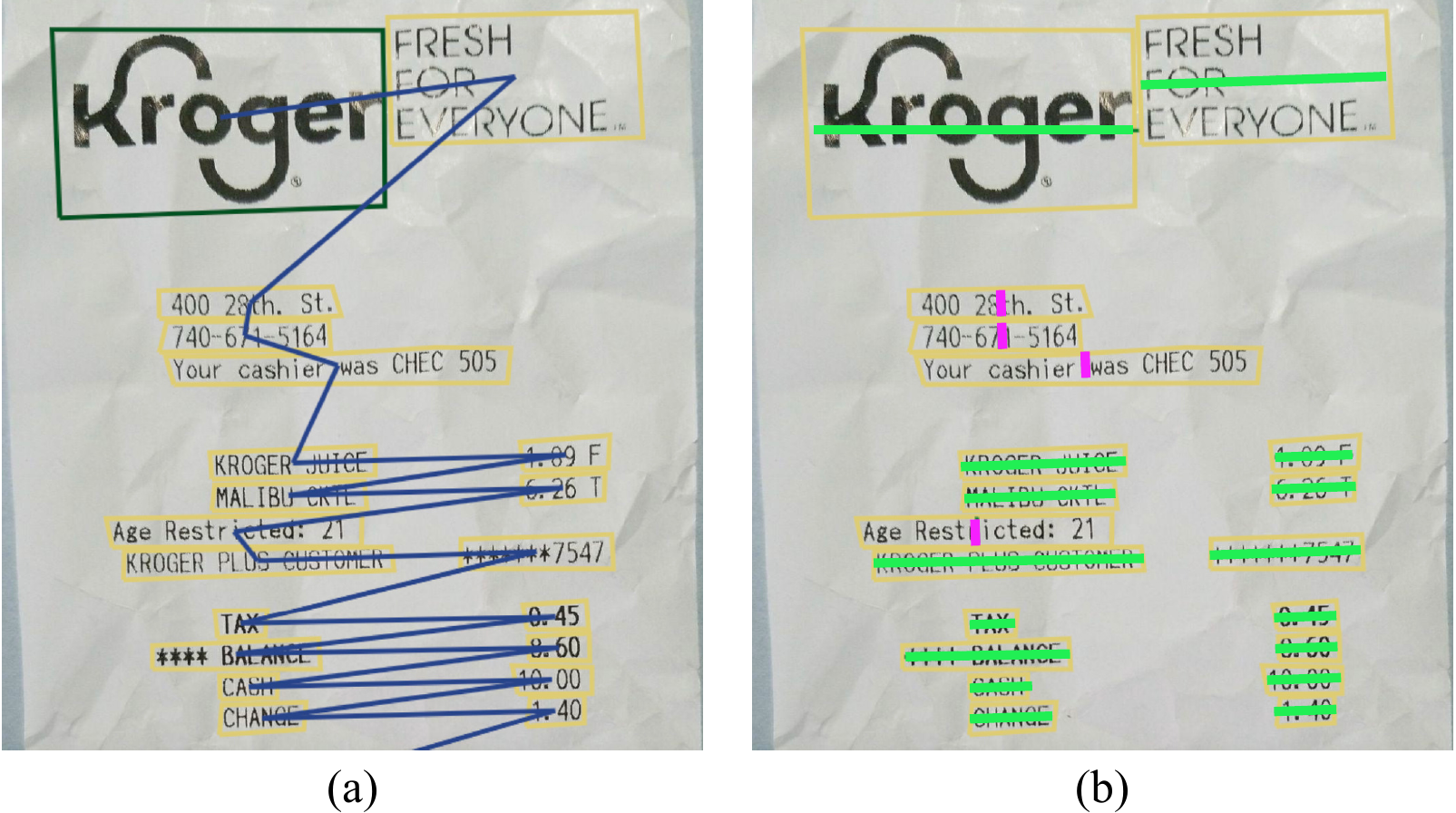}
\vspace{-3mm}
\centering
\caption{Labeling reading order patterns from annotations. (a) Ground truth from human annotated paragraphs and reading order. (b) Reading order pattern inferred from the annotated sequence --- column-wise indicated by a vertical/purple line in each paragraph and row-wise by a horizontal/green line. }
\label{fig:7}       % Give a unique label
\end{figure}

We prepared a dataset with human annotated layout data, including paragraphs as polygons and reading order groups where each group is an ordered sequence of paragraphs. Figure \ref{fig:7} (a) shows a set of paragraphs, where the reading order starts with the green paragraph and follows the jagged line.

\begin{table}[b!]
\centering

\noindent\begin{tabular}{p{0.95\textwidth}}
\noindent \hrulefill
\vspace{-5mm}

\noindent \hrulefill

\noindent \textbf{Algorithm 3: Pattern Labeling from Annotated Reading Order}

\noindent Input: A sequence of ground truth paragraphs $p_1, p_2, p_3, \cdots, p_n$ represented as

\hspace{9mm} rectangular boxes.

\vspace{-1mm}
\noindent \hrulefill
\vspace{-3mm}

\begin{enumerate}

\item Between each consecutive pair of paragraphs $(p_i, p_{i+1})$, we categorize their geometrical relation $R_{i,i+1}$ as one of $\{\mathit{vertical}, \mathit{horizontal}, \mathit{unknown}\}$.

\begin{enumerate}

\item Calculate $\alpha$, the circular mean angle of the two boxes' rotation angles.

\item Rotate the boxes of $p_i$ and $p_{i+1}$ around (0, 0) by $-\alpha$, denoted as $b_i$ and $b_{i+1}$.

\item Axis aligned box $c$ is the minimum containing box of both $b_i$ and $b_{i+1}$.

\item if $y_{\mathit{overlap}}(b_i, b_{i+1}) < 0.1\cdot\mathit{height}(c)\;\mathrm{and}\;
y_{\mathit{center}}(b_i) < y_{\mathit{center}}(b_{i+1})$

\hspace{5mm} $R_{i,i+1} = \mathit{vertical}$

\item else if $x_\mathit{overlap}(b_i, b_{i+1}) < 0.1\cdot\mathit{width}(c)\; \mathrm{and} \; x_\mathit{center}(b_i) < x_\mathit{center}(b_{i+1})$

\hspace{5mm} if $c$ does not cover paragraphs other than $p_i$, $p_{i+1}$

\hspace{10mm} $R_{i,i+1} = \mathit{horizontal}$ \hspace{10mm} \# mostly tabular structures

\hspace{5mm} else

\hspace{10mm} $R_{i,i+1} = \mathit{vertical}$ \hspace{14mm} \# mostly multi-column text

\item In other conditions, $R_{i,i+1} = \mathit{unknown}$

\end{enumerate}

\item Decide the reading order pattern for paragraph $p_i$ from $R_{i-1,i}$ and $R_{i,i+1}$.

\begin{enumerate}
\item $(\mathit{unknown}, \mathit{unknown}) \rightarrow \mathit{unknown}$

\item In case of one unknown, the other one decides the pattern:  $\mathit{vertical} \rightarrow$ column-wise, $\mathit{horizontal} \rightarrow$ row-wise,.

\item If neither is unknown, $(\mathit{vertical}, \mathit{vertical}) \rightarrow$ column-wise, otherwise it is row-wise.

\end{enumerate}
\end{enumerate}

\vspace{-4mm}
\noindent \hrulefill
\end{tabular}
\vspace{1mm}
\end{table}

While the edge clustering labels are straightforward from the paragraph polygons, the reading order pattern labeling is less trivial because we need to derive binary labels from ground truths of paragraph ordering. We decide the pattern of a paragraph by comparing its position with its predecessor and successor. Figure \ref{fig:7} (b) shows an example, and detailed logic is elaborated in Algorithm 3.

\subsection{Limitations}

The node-edge classification model can produce reasonable reading order in most cases, but may fail for complex layouts with multiple tabular sections placed closely, like the cross section errors in Figure \ref{fig:8} (a). The root cause is the lack of higher level layout structure parsing with the two classification tasks. Data annotation at section level is generally hard because there is no universal agreement on the exact definition of sections among text. Figure \ref{fig:8} (b) shows the result with extra section level clustering trained on a domain specific dataset. There is significant improvement, yet cross domain generalization is not guaranteed, and we can still see imperfections in the multi-section reading order due to section prediction errors.

\begin{figure}
\includegraphics[height=0.95\textwidth,angle=90]{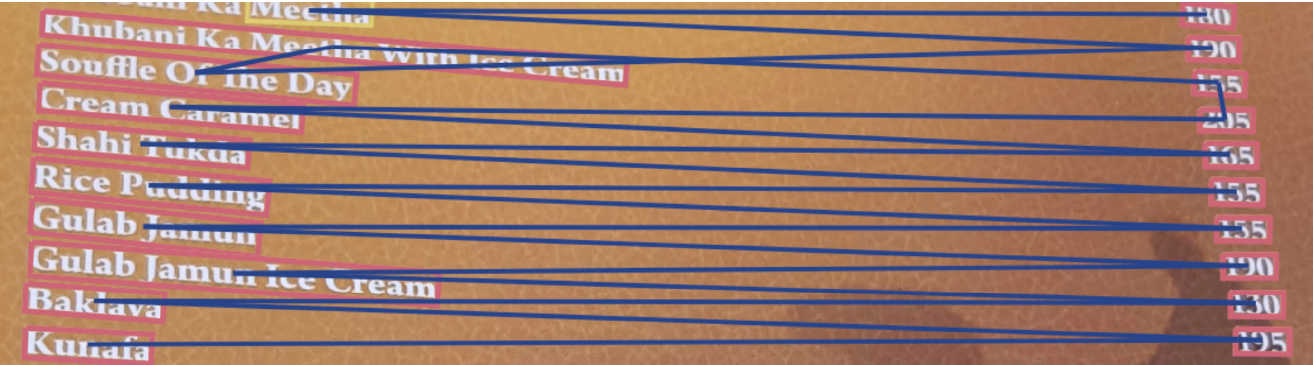}
\vspace{-2mm}
\centering
\caption{ Cluster-and-sort result on the cropped menu from Fig. \ref{fig:1} (a). Although the model correctly predicts the row-wise pattern, reading order is still incorrect due to the perspective distortion and the unusually large spacing between the two columns. }
\label{fig:9}       % Give a unique label
\end{figure}

Another limitation is that our model is not a reliable source for parsing table structures like \cite{table2022}. Figure \ref{fig:9} shows the reading order result of the image in Figure \ref{fig:1} (a). Note that in the sorting algorithm, we rotate all the bounding boxes to zero out their mean angle. But when the boxes are at different angles due to distortions, there will still be slanted line boxes and misaligned table rows after all the rotations, so the topological sort on the axis-aligned containing boxes cannot guarantee the right order. In presence of tables, a separate model with structure predictions will likely perform better.

\section{Experiments}

We experiment with the GCN model with predictions on reading order pattern and paragraph clustering, together with the cluster-and-sort algorithms. 

\subsection{Datasets and Evaluation Metrics}

Various metrics have been used to evaluate reading order, such as Spearman's footrule distance, Kendall's Tau rank distance used in \cite{readingorderhand2022} and BLEU scores in \cite{e2esequence2020}. These metrics can accurately measure order mismatches, but also require full length ground truth order for comparison.

We created an annotated layout dataset where reading order ground truths are partially annotated, i.e. some subsets of paragraphs form reading order groups with annotated order, and the order among groups is undefined. This makes it more flexible to match realistic user requirements and less suitable for full ranking metrics. So instead, we use a normalized Levenshtein distance \cite{lev1966} which measures the minimum number of word operations (insertions and deletions) needed to equalize two lists. For each reading order group, we take the ordered list of paragraphs and find all the OCR words $W$ contained in these polygons. The word order within each paragraph is taken directly from OCR (mostly accurate for a single paragraph). Then we find the shortest subsequence of the serialized OCR output that contains all the words in $W$, compute its Levenshtein distance to $W$, and multiply it by the normalization factor $1/\vert W \vert$. 

Besides our annotated set, we test the model with PubLayNet \cite{publaynet2019} because of its variety on layout components with different reading order patterns. Although there is no ground truth of reading order, we take ``text'' instances as paragraphs with column-wise pattern, and ``table''/``figure'' types as containers of text lines with row-wise pattern. Thus, we are able to train the same multi-task GCN model.
The annotated set contains 25K text images in English for training and a few hundred test images for each of the available languages, and PubLayNet contains 340K training images and 12K validation images all in English.

\begin{figure}[t]
\includegraphics[height=\textwidth,angle=90]{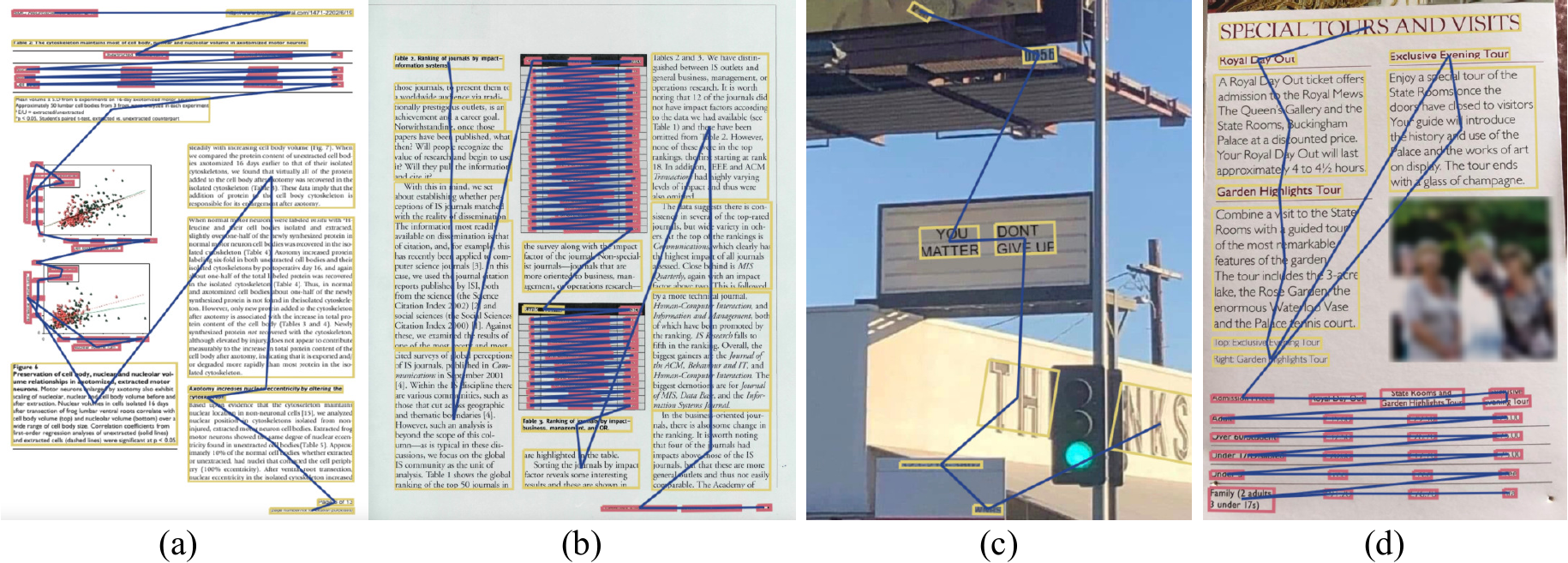}
\vspace{-7mm}
\centering
\caption{Reading order results from (a) PubLayNet \cite{publaynet2019}, (b) PRIMA Contemporary dataset \cite{layout2009}, (c) the ambiguous example from Figure \ref{fig:1} with a positive interpretation, and (d) our evaluation set.}
\label{fig:10}       % Give a unique label
\vspace{-2mm}
\end{figure}

\subsection{Model Setup}
 
The model is built as shown in Figure \ref{fig:4}, with the OCR engine from Google Cloud Vision API producing text lines and their spatial features. Edge image features are from a bi-linear interpolation on the MobileNetV3 output with $16\times3$ points each box and dropout rate 0.5. The TF-GNN~\cite{tfgnn} based GCN backbone uses 10 steps of weight-sharing graph convolutions, with node feature dimension 32 and message passing hidden dimension 128. Edge-to-node pooling uses a 4-head attention with 3 hidden layers of size 16 and dropout rate 0.5. Total number of parameters is 267K including 144K from MobileNetV3-Small.

We train the model for 10M steps with randomized augmentations including rotation and scaling, so the model can adapt to a full range of inputs. The OCR boxes are transformed together with the image in each training example, resulting in better robustness than previous approaches (Figure~\ref{fig:6}).

\vspace{-2mm}

\subsection{Baselines}
\label{sec:baselines}

Most commercial OCR systems use a topological sort like in \cite{bidim2003} with one of the two patterns. We use column-wise pattern in the basic baseline as it produces better scores than row-wise in our evaluations, and is close to the default output order from the OCR engine we use.

\begin{figure}[t]
\includegraphics[height=\textwidth,angle=90]{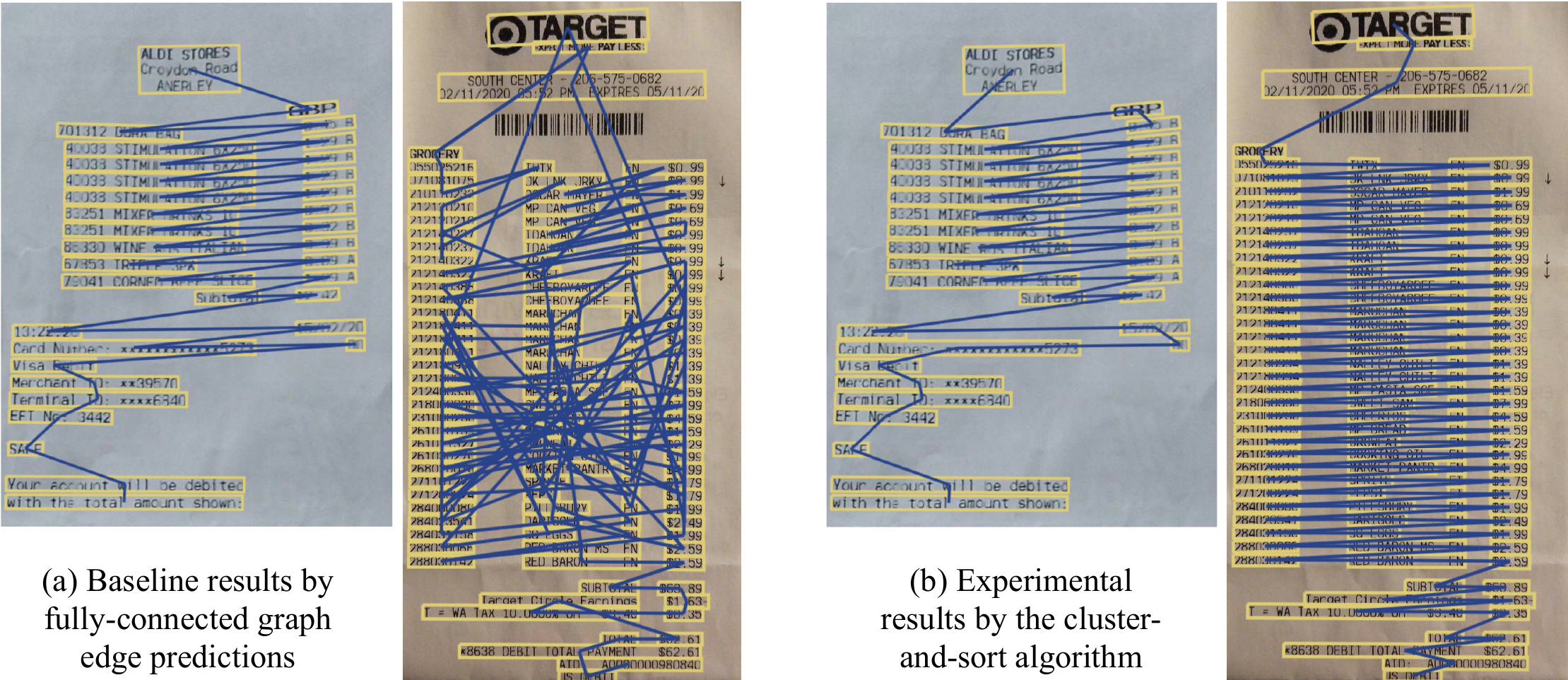}
\vspace{-6mm}
\centering
\caption{Comparison between the fully-connected graph model and our approach on two receipt examples. The full graph predictions perform well on the sparse example, but fail on the dense one. }
\label{fig:11}       % Give a unique label
\end{figure}

In addition, we implement a GCN model that directly predicts edge directions on a fully connected graph similar to the model in \cite{e2esequence2020}. Figure~\ref{fig:11} shows two examples with comparison between this baseline and our approach, with supports
the scalability discussion in subsection \ref{sec:related_reading_order}.

\vspace{-2mm}

\subsection{Results}

\begin{table}[b!]
%\footnotesize
\centering
\caption{Scores of the two classification tasks on PubLayNet and our labelled paragraph reading order dataset.}
\vspace{-1mm}
\label{tab:1}       % Give a unique label
\begin{tabular}{ccccccc}
\hline \hline\noalign{\smallskip}
Dataset & \multicolumn{3}{l}{Reading order pattern} \ \ \ & \multicolumn{3}{l}{Paragraph clustering \ \ \ \ \ } \\
& Precision & Recall & F1 & Precision & Recall & F1 \\
\noalign{\smallskip}\hline\noalign{\smallskip}
PubLayNet & 0.998 & 0.995 & 0.997 & 0.994 & 0.996 & 0.995 \\
Annotated ordered paragraphs \ \ & 0.828 & 0.805 & 0.819 & 0.895 & 0.909 & 0.902 \\
\noalign{\smallskip}\hline
\end{tabular}
\vspace{6mm}
%\footnotesize
\centering
\caption{F1 scores from the image feature ablation test.}
\vspace{-1mm}
\label{tab:2}       % Give a unique label
\begin{tabular}{ccccccc}
\hline \hline\noalign{\smallskip}
Boxes for image & \multicolumn{3}{l}{Reading order pattern} \ \ \ & \multicolumn{3}{l}{Paragraph clustering \ \ \ \ \ } \\
feature RoI pooling \ \ & Precision & Recall & F1 & Precision & Recall & F1 \\
\noalign{\smallskip}\hline\noalign{\smallskip}
n/a & 0.800 & 0.803 & 0.802 & 0.887 & 0.895 & 0.891 \\
Node boxes & 0.819 & 0.781 & 0.800 & 0.870 & 0.903 & 0.886 \\
Edge boxes & 0.828 & 0.805 & 0.819 & 0.895 & 0.909 & 0.902 \\
\noalign{\smallskip}\hline
\end{tabular}
\end{table}

We train the multi-task model with PubLayNet and our paragraph reading order set added with the menu photos labelled from human annotations. From Table~\ref{tab:1}, we can see the difference in the difficulty between the two sets. Real-world images from our dataset have much larger variations on layout styles and image degradations that make the same tasks much harder to learn.

We also test the effectiveness of the edge box RoI pooling by an image feature ablation test, where the baseline is the model with all image features removed, compared against ones with node box RoI pooling and edge box RoI pooling. Table \ref{tab:2} shows that node box RoI does not help at all, even with a slight accuracy drop compared with the baseline. These results confirm our previous hypothesis that the image backbone mainly helps the model by discovering visual cues out of text bounding boxes, and edge boxes are much more effective for this purpose.

\begin{table}[b!]
%\footnotesize
\centering
\caption{Normalized Levenshtein distance (lower is better) on a multi-language reading order evaluation set. Training data is only available in English. }
\vspace{-1mm}
\label{tab:3}       % Give a unique label
\begin{tabular}{l|c|r|c|c|c}
\hline \hline\noalign{\smallskip}
Language \ & \ Training \ & \ Test set \ & \ All-column-wise \ & \ Fully-connected \ & 2-task GCN \\
 & set size & \multicolumn{1}{c|}{size} & baseline & graph baseline & cluster-and-sort \\
\noalign{\smallskip}\hline\noalign{\smallskip}
English & 25K & 261 \hspace{4pt} & 0.146 & 0.126 & 0.098 \\
\cline{2-2}
French & \multirow{7}{*}{n/a} & 218 \hspace{4pt} & 0.184 & 0.144 & 0.119 \\
Italian & & 189 \hspace{4pt} & 0.172 & 0.145 & 0.122 \\
German & & 196 \hspace{4pt} & 0.186 & 0.162 & 0.112 \\
Spanish & & 200 \hspace{4pt} & 0.183 & 0.103 & 0.097 \\
Russian & & 1003 \hspace{4pt} & 0.202 & 0.159 & 0.148 \\
Hindi & & 990 \hspace{4pt} & 0.221 & 0.181 & 0.152 \\
Thai & & 951 \hspace{4pt} & 0.131 & 0.111 & 0.104 \\
\noalign{\smallskip}\hline
\end{tabular}
\end{table}

Finally, we measure the normalized Levenshtein distance for reading order produced by the GCN and the cluster-and-sort algorithm, and compare it against the two baseline methods in subsection \ref{sec:baselines}. As in Table \ref{tab:3}, our algorithm can greatly improve reading order quality across all Latin languages, even though the training data is only available in English. The model also works well for examples out of our datasets. Figure \ref{fig:10} includes images from various sources, demonstrating the effectiveness of our model with inputs ranging from digital/scanned documents to scene images.

\vspace{-2mm}

\section{Conclusions and Future Work}

We show that GCN is highly efficient at predicting reading order patterns and various layout segmentation tasks, which is further enhanced with a small image backbone providing edge RoI pooled signals. Our model is small in size and generalizes well enough to be deployable on any platform to improve OCR quality or downstream applications.

\begin{figure}[t]
\includegraphics[height=\textwidth,angle=90]{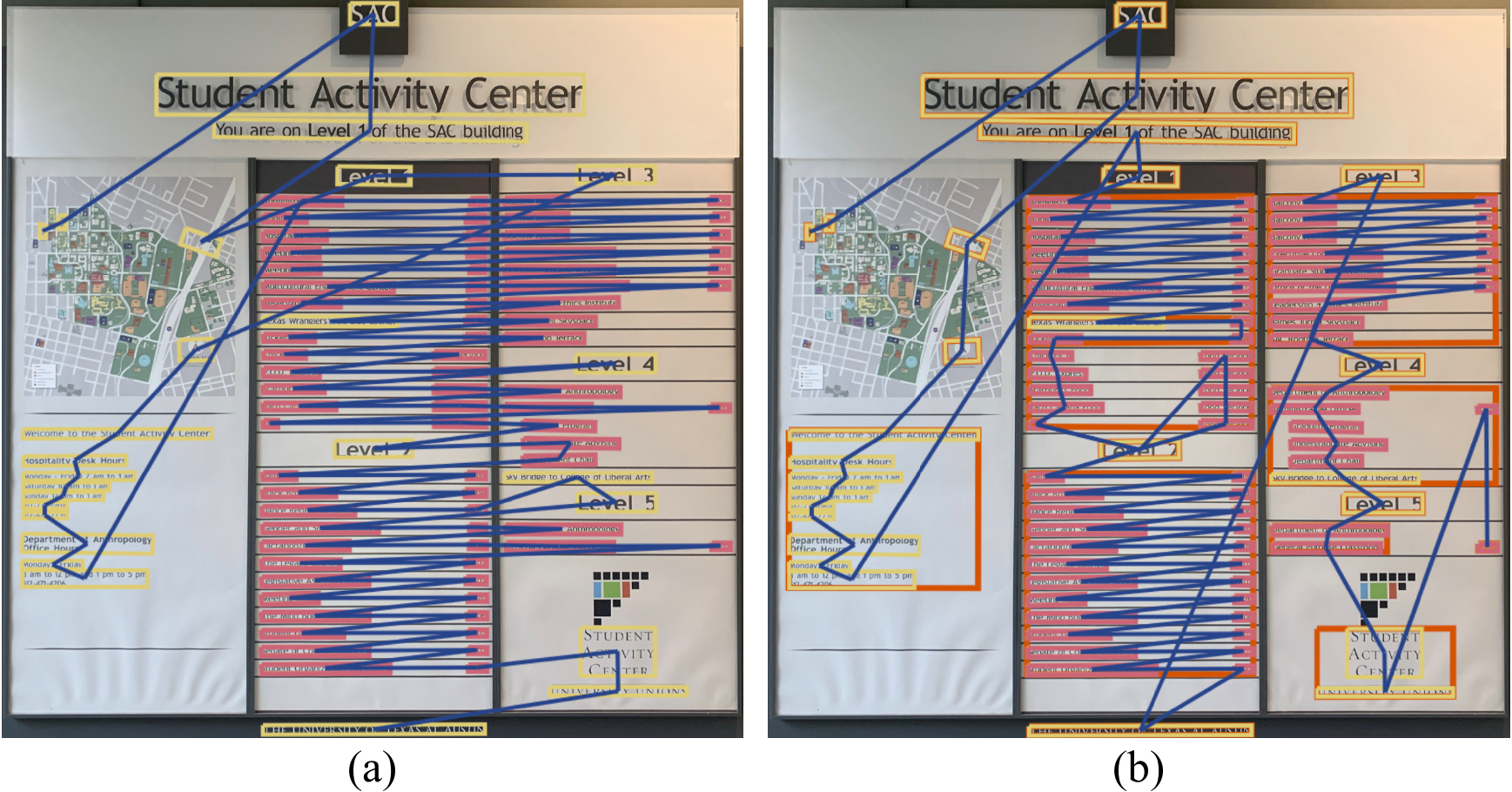}
\vspace{-5mm}
\centering
\caption{A multi-section example. (a) Paragraphs with row-wise pattern are clustered into overly large regions, causing incorrect cross-section reading order. (b) With section level clusters (shown in orange) added into Algorithm 1, multi-table results can be improved. }
\vspace{-5mm}
\label{fig:8}       % Give a unique label
\end{figure}

In addition, the GCN model has the potential to handle more than two tasks. We tried an extra edge prediction task trained with a dataset of menu photos with section level polygon annotations. Unlike general document or scene text images, menus like Figure \ref{fig:3} usually have clearly defined sections like main dishes, side dishes, drinks, etc. Therefore, the menu dataset has accurate and consistent section level ground truth for model training. The 3-task GCN model provides higher-level layout information to the clustering algorithm and helps produce Figure \ref{fig:8} (b), a major improvement on reading order. Still, there is domain specific knowledge on menu sections that does not always generalize well. And because most evaluation examples have relatively simple layouts, the 3-task model has not produced better results than the 2-task model in our experiments. Nevertheless, we think section level ground truth or higher-level layout structural information will be valuable for further reading order improvements. Future work will explore the possibilities of both data and modeling approaches for parsing layout structures.

\subsubsection{Acknowledgements} The authors would like to thank Ashok C. Popat and Chen-Yu Lee for their valuable reviews and feedback.
\bibliographystyle{splncs04}

\end{document}